\documentclass{article}
\usepackage{spconf,amsmath,graphicx,hyperref}

\usepackage[utf8]{inputenc} 
\usepackage[T1]{fontenc}    
\usepackage{url}            
\usepackage{booktabs}       
\usepackage{amsfonts}       
\usepackage{nicefrac}       
\usepackage{microtype}      
\usepackage{xcolor}         
\usepackage{multirow}
\usepackage{multicol}
\usepackage{tcolorbox}
\usepackage{pifont}
\usepackage{graphicx}
\usepackage{inconsolata}
\usepackage{xspace}
\usepackage{latexsym}
\usepackage{utfsym}
\usepackage{algorithm}
\usepackage{algorithmic}
\usepackage{array}
\usepackage{colortbl}
\usepackage{amsmath}

\definecolor{gbypink}{rgb}{0.99, 0.91, 0.95} 
\definecolor{lightblue}{rgb}{0.81, 0.94, 1.0}

\newcommand{\hcmark}{\ding{52}\rotatebox[origin=c]{-9.2}{\kern-0.7em\ding{55}}}

\definecolor{mygreen}{RGB}{0,128,0}  
\definecolor{myred}{RGB}{255,0,0}

\title{Improving the Reasoning of Multi-Image Grounding in MLLMs via Reinforcement Learning}
\name{Bob Zhang$^{1*}$ ,  Haoran Li$^{1,2*}$ ,  Tao Zhang$^{3}$ , Jianan Li$^{4}$ ,  Cilin Yan$^{1}$ ,   Xikai Liu$^{1}$ ,  Jiayin Cai$^{1}$  , Yanbin Hao$^{5}$}
  
\address{$^1$Xiaohongshu Inc. \\
$^2$University of Science and Technology of China \\ 
$^3$Wuhan University \\
$^4$Technical University of Munich \\
$^5$Hefei University of Technology }

\begin{document}
%
\maketitle
%
\begin{abstract} 

Multimodal Large Language Models (MLLMs) perform well in single-image visual grounding but struggle with real-world tasks that demand cross-image reasoning and multi-modal instructions. To address this, we adopt a reinforcement learning (RL) based post-training strategy for MLLMs in multi-image grounding tasks. We first synthesize high-quality chain-of-thought (CoT) data for cold-start initialization, followed by supervised fine-tuning (SFT) using low-rank adaptation (LoRA). Subsequently, we apply rejection sampling with the merged SFT model to curate reliable RL data and use rule-based RL to guide the model toward optimal reasoning paths. Extensive experiments demonstrate the effectiveness of our approach, achieving +9.04\% on MIG-Bench and +4.41\% on average across seven out-of-domain benchmarks.
\end{abstract}


\renewcommand{\thefootnote}{}
\footnotetext{$^*$Equal contributions.}
\renewcommand{\thefootnote}{\arabic{footnote}}

\vspace{-0.3cm}
\section{Introduction}
\begin{figure}[t]
  \centering
  \includegraphics[width=\columnwidth]{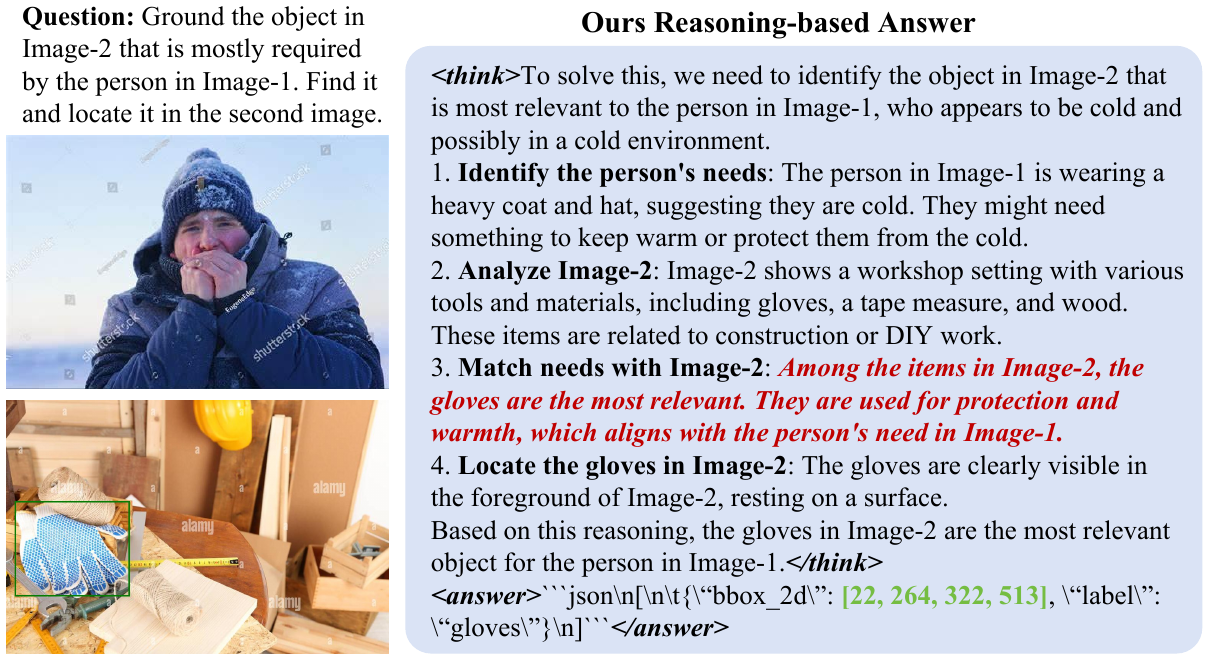}
  \vspace{-0.6cm}
  \caption{Example of our multi-image grounding results.}
  \label{fig:teasor}
\vspace{-0.6cm}
\end{figure}

\vspace{-0.2cm}
Traditional visual grounding localizes targets in a single image from simple language descriptions, where large language models (LLMs) have significantly improved performance. However, real-world applications require multi-image grounding with complex instructions and cross-image reasoning. As illustrated in Figure \ref{fig:teasor}(a), practical scenarios often require identifying the region that best matches the query image semantics. 

Recently, Migician~\cite{li2025migician} introduces a multi-image grounding dataset and employs a two-stage supervised fine-tuning (SFT) pipeline. However, their SFT approach primarily "memorizes"~\cite{chu2025sft}  understanding patterns and instruction-following behaviors in multi-image scenarios, rather than cultivating reasoning, which limits generalization to real-world scenarios.

Inspired by the recent success of reinforcement learning (RL)–based post-training frameworks for Large Reasoning Models (LRM)~\cite{guo2025deepseek}, we explore the potential of the RL paradigm to improve MLLMs' reasoning abilities in multi-image grounding scenarios. Our preliminary attempts applied RL directly to the open-source MLLM Qwen2.5-VL-7B~\cite{bai2025qwen2}. However, we observed frequent failures to generate correct responses during RL training, mainly due to the base model’s limited capacity to handle complex cross-image queries.

Building upon these foundations, we explore the use of RL to improve multi-image reasoning in MLLMs. Specifically, we propose a two-stage framework consisting of cold-start CoT-SFT initialization and rule-based RL training. First, we synthesize high-quality chain-of-thought (CoT) data using Qwen2.5-VL-72B and perform supervised fine-tuning with LoRA for cold-start initialization, equipping the model with basic multi-image reasoning ability. Next, we adopt a rule-based RL paradigm based on Group Relative Policy Optimization (GRPO)~\cite{shao2024deepseekmath} to guide the model toward discovering correct reasoning paths and improving generalization. However, we observe that a substantial portion of the data is of low quality. This prevents the model from generating response candidates with sufficient relative advantage gaps, thereby yielding limited supervisory signals for RL. To improve training efficiency, we apply rejection sampling based on predictions from the merged cold-start SFT model, to filter out such uninformative samples before the RL stage.

\begin{figure*}
  \centering
  \includegraphics[width=0.8\textwidth]{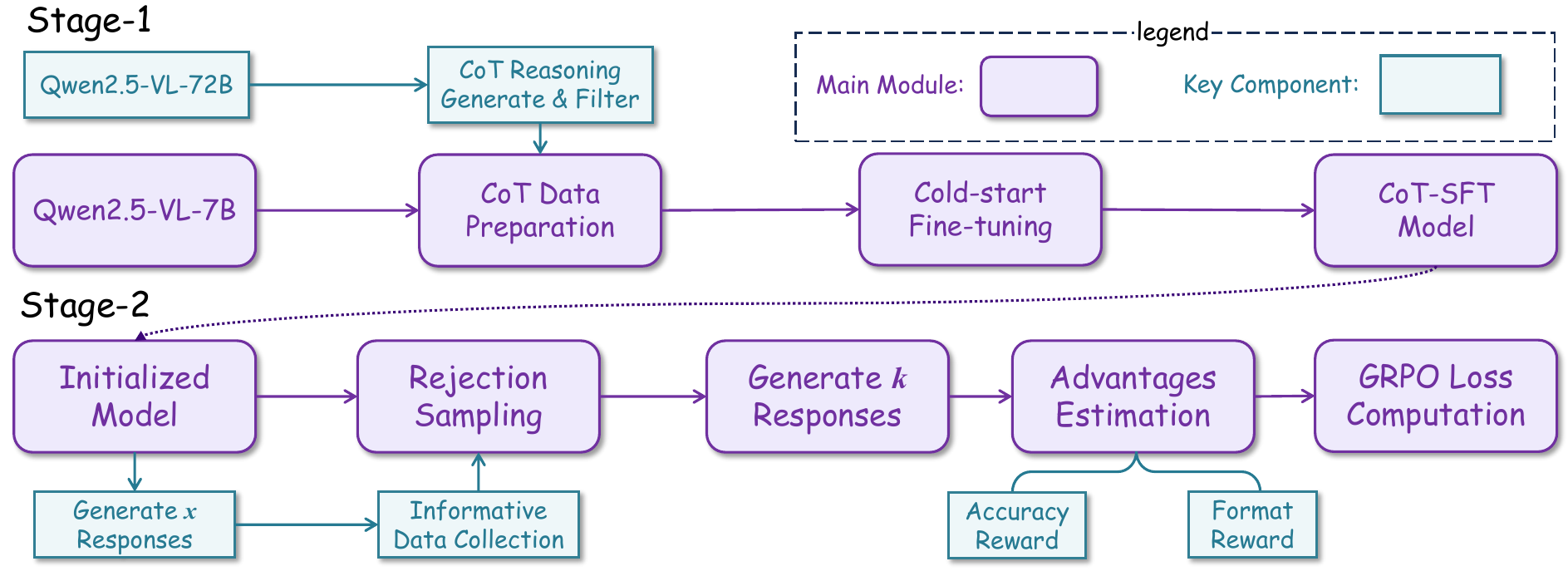}
  \vspace{-0.2cm}
  \caption{\textbf{The overview of our two-stage training paradigm}. Our training paradigm consists of a cold-start CoT-SFT initialization (stage-1) and a rule-based RL training (stage-2).}
  \label{fig:overview}
  \vspace{-0.2cm}
\end{figure*} 

To validate our approach, we conduct comprehensive experiments across multiple benchmarks. On the multi-image grounding benchmark MIG-Bench, our model achieves a gain of +9.04\% over the SFT baseline.  Moreover, we evaluate performance across seven diverse datasets covering zero-shot grounding, object detection and multi-image understanding. Overall, our model delivers the highest average performance, surpassing the SFT baseline by +4.41\%.

\vspace{-0.3cm}
\section{Method} \label{sec:3_method}
\vspace{-0.2cm}

\subsection{Preliminaries}
\vspace{-0.1cm}
We formally define the multi-image grounding task as follows: given a natural language description $t$, query images $Q$, and a set of $m$ target images ${I_1, I_2, ..., I_m}$, the model $M$ localizes a region of interest by generating a bounding box $G$ that satisfies the semantic and contextual constraints specified by $t$ and $Q$

Unlike traditional grounding with fixed instruction templates, multi-image grounding requires handling dynamic, context-sensitive instructions, which demand stronger comprehension and reasoning from MLLMs. To address this gap, we employ an RL-based training paradigm, building on recent advances in LRMs for complex reasoning. We select Qwen2.5-VL-7B for its strong multimodal understanding. As shown in Figure~\ref{fig:overview}, our training consists of two stages: cold-start CoT-SFT initialization and rule-based RL training, which are detailed in the following subsections.

\vspace{-0.2cm}
\subsection{Cold-start CoT-SFT Initialization} \label{sec:3_1_cold_start}

As observed in DeepSeek-R1~\cite{guo2025deepseek}, directly applying reinforcement learning from scratch may result in unstable convergence. 
The base model particularly struggles with complex multi-image contexts and accurate localization. To address this, we construct a high-quality CoT cold-start dataset to equip the MLLM with task-specific reasoning ability and boost initial performance before RL training.

\vspace{-0.4cm}
\subsubsection{CoT Reasoning Dataset} 
\vspace{-0.1cm}

Our CoT reasoning dataset is constructed on top of MGrounding-630k dataset~\cite{li2025migician}. 
To maintain the model’s capability in single-image grounding, we augment it with a small portion of single-image grounding data from RefCOCO/+/g~\cite{mao2016generation, yu2016modeling} and object detection datasets, ODINW~\cite{li2022grounded} and V3Det~\cite{wang2023v3det}. This hybrid approach enables comprehensive
training across both multi-image and single-image grounding tasks.

We employ Qwen2.5-VL-72B for generating CoT reasoning data. Following the paradigm of DeepSeek-R1, our generation process takes a question-image pair accompanied by a fixed instructional prompt as input. The MLLM then produces structured CoT rationales in the following format:

\texttt{\{Question\} First output the thinking process in <think> </think> tags and then output the final answer in <answer> </answer> tags. Output the bounding box coordinates in JSON format.}

To ensure high-quality training data, we apply an additional filtering strategy. For each sample, we prompt the Qwen2.5-VL-72B model to generate four distinct responses. Next, we assess sample consistency by computing the Intersection over Union (IoU) accuracy across all responses and keep only those with perfect agreement (IoU accuracy = 1.0).
This process yields 56k high-confidence CoT cold-start samples, including 54k from MGrounding-630k, 1k from RefCOCO/+/g, and 1k from ODINW and V3DET.

\vspace{-0.4cm}
\subsubsection{Cold-start Supervised Fine-tuning} 
\vspace{-0.1cm}

Leveraging our curated CoT dataset, we initialize Qwen2.5-VL-7B with multi-image comprehension and reasoning. We apply strong supervision via next-token prediction across the generation process, enabling the model to learn both reasoning traces and spatial grounding.

Since full-parameter fine-tuning can reduce reasoning flexibility and weaken generalization and language quality, we adopt LoRA, updating only low-rank matrices. This approach preserves core knowledge while enhancing perception and reasoning for multi-image grounding. The trained LoRA modules are then merged with the base MLLM, producing our Stage-1 CoT-SFT model with enhanced capabilities.

\subsection{Rule-based RL Training}

In the second stage, we implement group relative policy optimization (GRPO)~\cite{shao2024deepseekmath}, a rule-based RL method that refines policy updates using group-wise relative constraints. To improve training efficiency, we integrate rejection sampling before GRPO, filtering out uninformative samples to accelerate convergence and stabilize learning.

\vspace{-0.4cm}
\subsubsection{Group Relative Policy Optimization}
\vspace{-0.1cm}

The MLLM generates a group of $G$ complete responses $\{o_1, o_2,... , o_G\}$ with current policy $\pi_\theta$. The output response is required to contain a full CoT reasoning process and a final bounding box prediction. For each response $o_i$, we compute a scalar reward $r_i$, and normalize these rewards to estimate its group-relative advantage $A_i$, formally as:
\vspace{-0.1cm}
\begin{equation}
    A_i = \frac{r_i - mean(\{r_j\}_{j=1}^N)}{std(\{r_j\}_{j=1}^N)}
\end{equation}
\vspace{-0.15cm}
Then, the GRPO training objective can be defined as follows:
\vspace{-0.15cm}
\begin{align}
&\mathcal{J} _{GRPO}(\theta )=\frac{1}{N}\sum_{i=1}^N \nonumber \\
&[ \min ( \frac{\pi _{\theta}(o_i|q)}{\pi _{\theta _{old}}(o_i|q)}A_i,\mathrm{clip}( \frac{\pi _{\theta}(o_i|q)}{\pi _{\theta _{old}}(o_i|q)},1-\epsilon ,1+\epsilon ) A_i ) \nonumber \\
&-\beta ·\mathcal{K} \mathcal{L} (\pi _{\theta}(o_i|q)|\pi _{ref}(o_i|q)],
\label{eq: objective}
\end{align}
where $\pi _{\theta _{old}}$ is the previous policy before update, and $\pi_ {ref}$ is the fixed CoT-SFT policy after stage-1 training.
The hyperparameters $\epsilon$ and $\beta$ control the policy update clipping range and KL divergence penalty strength, respectively.

\begin{table*}[t]
\centering
\resizebox{0.8\textwidth}{!}{
\begin{tabular}{l|ccccccc | ccc|c}
\toprule
\multirow{4}{*}{\textbf{Models}} & \multicolumn{7}{|c|}{\textbf{ Referential Grounding }} & \multicolumn{3}{c}{\textbf{Spontaneous Grounding }} & \multirow{4}{*}{\textbf{AVG}} \\ 
\cmidrule(lr){2-8} \cmidrule(lr){9-11} 

\multicolumn{1}{c}{} & \multicolumn{4}{c|}{\textbf{Visual Reference}} & \textbf{Textual} & \multicolumn{2}{|c|}{\textbf{Visual+Textual}} & \multicolumn{2}{|c|}{\textbf{Difference}} & \textbf{Similarity} & \\ 
\cmidrule(lr){2-5} \cmidrule(lr){6-6} \cmidrule(lr){7-8} \cmidrule(lr){9-10} \cmidrule(lr){11-11}

\multicolumn{1}{c|}{} & \textbf{OT} & \textbf{MV} & \textbf{Region} & \textbf{Refer} &  \multicolumn{1}{|c|}{\textbf{GG}} & \textbf{Reason} & \multicolumn{1}{c|}{\textbf{Co-Re}} & \textbf{Static} & \textbf{Robust} & \multicolumn{1}{|c|}{\textbf{Common}} & \\

\midrule

\rowcolor{gray!20} \multicolumn{12}{c}{\textbf{70B-Scale MLLMs}}\\
\midrule

LLaVA-OV-72B & 12.91 & 7.64 & 2.14 & 17.83 & 21.60 & 11.88 & 8.55 & 13.26 & 5.34 & 26.84 & 13.65 \\
InternVL3-78B & 27.08 & 14.58 & 10.44 & 50.51 & 38.08 & 45.54 & 17.09 & 10.04 & 9.57 & 24.12 & 24.71 \\
Qwen2.5-VL-72B & 34.32 & 29.17 & 8.31 & 62.63 & 59.92 & \underline{66.34} & \textbf{41.03} & 43.75 & \textbf{46.81} & 69.98 & 46.23 \\
\midrule 

\rowcolor{gray!20} \multicolumn{12}{c}{\textbf{7B-Scale MLLMs}}\\
\midrule
LLaVA-OV-7B & 0.18 & 1.04 & 1.08 & 9.09 & 15.43 & 6.93 & 0.85 & 6.06 & 3.19 & 3.43 & 4.73 \\
Minicpm2.6 & 9.82 & 6.25 & 1.75 & 11.11 & 10.02 & 2.97 & 2.56 & 14.58 & 2.13 & 14.34 & 7.55 \\
mPLUG-Owl3 & 8.55 & 7.64 & 2.41 & 7.07 & 22.85 & 9.09 & 5.98 & 18.56 & 6.38 & 34.93 & 12.35 \\
InternVL3-8B & 14.84 & 6.94 & 12.13 & 7.07 & 34.87 & 16.83 & 2.56 & 23.67 & 14.89 & 47.99 & 18.18 \\
\midrule
Qwen2.5-VL-7B & 15.23 & 5.56 & 4.07 & 13.13 & 32.26 & 3.96 & 2.56 & 28.03 & 5.32 & 21.81 & 13.19 \\
+ SFT  & 26.59 & \underline{35.07} & \underline{39.32} & 79.80 & 59.72 & 53.47 & 23.93 & \textbf{53.22} & 44.68 & \underline{82.11} & 49.79 \\
+ CoT-SFT & \underline{46.14} & 34.03 & 28.93 & \underline{80.81} & \underline{63.93} & \underline{66.34} & 34.19 & 47.35 & \underline{45.74} & 76.59 & \underline{52.41} \\
\rowcolor{lightblue} \textbf{Ours} & \textbf{63.41} & \textbf{38.54} & \textbf{51.45} & \textbf{82.83} & \textbf{69.54} & \textbf{67.33} & \underline{37.61} & \underline{51.52} & 43.62 & \textbf{82.48} & \textbf{58.83} \\
\bottomrule
\end{tabular}
}
\vspace{-0.2cm}
\caption{\textbf{Performance comparison on MIG-Bench.} OT, MV, GG, and Co-Re denote object tracking, multi-view grounding, group grounding, and correspondence. Values with * are based on 20\% sampled test examples for human evaluation. The best and second-best results are shown in \textbf{bold} and \underline{underline}, respectively.}
\vspace{-0.2cm}
\label{tab:mig-bench}
\end{table*}

\vspace{-0.4cm}
\subsubsection{Rejection Sampling} 
\vspace{-0.1cm}

When generating n predictions per sample during rule-based RL training, batches that are uniformly correct or uniformly incorrect yield zero reward variance, resulting in a vanishing advantage signal and ineffective optimization. To address this issue, we perform rejection sampling between stage-1 SFT and stage-2 RL training using the fixed stage-1 CoT-SFT model. Specifically, we (1) generate multiple predictions per sample and (2) discard samples exhibiting either complete correctness or complete failure. Retaining only partially correct cases ensures non-zero reward variance and provides stronger learning signals for optimization. After filtering, we obtain a high-quality dataset of 174k samples for stage-2 training.

\vspace{-0.4cm}
\subsubsection{Reward Design} 
\vspace{-0.1cm}

We employ two reward functions for the multi-image grounding task: accuracy reward $r_{acc}$ and format reward $r_{format}$.

\textit{Accuracy Reward}. Given the ground truth bounding box $B$ and the predicted bounding box $\hat{B}$, we calculate the $IoU(B, \hat{B})$ as our accuracy reward, where $IoU$ denotes the Intersection over Union metric.

\textit{Format Reward}. We adopt a format reward similar to DeepSeek-R1, requiring the model output to follow the format: “<think>...</think> <answer>...</answer>”. The reward score is set to 1 only when the output adheres to the required format; otherwise, the score is 0. In addition, We ensure that bounding boxes are generated in valid JSON format.

The total reward is computed as a weighted sum of both the accuracy reward and the format reward:
\vspace{-0.1cm}
\begin{equation}
    r = \lambda_{acc} r_{acc} + \lambda_{format} r_{format}
\end{equation}
where $\lambda_{acc}$ and $\lambda_{format}$ denote the weights assigned to the accuracy reward and the format reward, respectively.
\vspace{-0.3cm}
\section{Experiments} 
\vspace{-0.2cm}

In this section, we describe the experimental setup (Section~\ref{sec:4_1_setup}), present comparison results (Section~\ref{sec:4_2_comparsion}), and conduct ablation studies (Section~\ref{sec:4_3_ablation}).

\begin{table*}[t] 
\centering
\resizebox{0.85\textwidth}{!}{
\begin{tabular}{l|c c| c c|c|c c|c}
\toprule
\multirow{2}{*}{\textbf{Model}} & \multicolumn{2}{c|}{\textbf{Multi-image Grounding}} & \multicolumn{2}{c|}{\textbf{Single-image Grounding}} & \multicolumn{1}{c|}{\textbf{Object detection}} & \multicolumn{2}{c|}{\textbf{Multi-image Understanding}} & \multirow{2}{*}{\textbf{AVG}} \\
\cmidrule(lr){2-3}  \cmidrule(lr){4-5} \cmidrule(lr){6-6} \cmidrule(lr){7-8}
& ReVOS~\cite{yan2024visa} & MC-Bench~\cite{xu2024mc} & LISA~\cite{lai2024lisa} & Ref-L4~\cite{chen2025revisiting} & ODINW~\cite{li2022grounded} & MMIU~\cite{meng2024mmiu} & BLINK~\cite{fu2024blink}  \\
\midrule
Qwen2.5-VL-7B  & 30.07 & 24.37 & 55.09 & 60.67 & 46.91 & 47.52 & 58.07 & 46.10 \\
+ SFT          & 46.33 & 52.12 & 54.64 & 69.03 & 49.82 & 48.21 & 56.56 & 53.82 \\
\rowcolor{lightblue} \textbf{Ours} & \textbf{49.16} & \textbf{58.49} & \textbf{59.44} & \textbf{75.26} & \textbf{54.15} & \textbf{49.93} & \textbf{61.18} & \textbf{58.23} \\
\bottomrule
\end{tabular}
}
\vspace{-0.2cm}
\caption{\textbf{Evaluation on several reasoning grounding, single-image grounding, object detection, and multi-image understanding benchmarks.} The best results are marked in \textbf{bold}.}
\vspace{-0.3cm}
\label{tab:all_bench}
\end{table*}

\vspace{-0.3cm}
\subsection{Experimental Setup} \label{sec:4_1_setup}
\vspace{-0.1cm}
\textbf{Implementation Details.} We use weighted rewards with $\lambda_{acc}=1.0$ and $\lambda_{format}=0.5$. Sampling uses a temperature of 0.7, a maximum length of 1024 tokens, and 8 rollouts per prompt. Training uses learning rates of $1\times10^{-4}$ (SFT) and $5\times10^{-5}$ (GRPO), with batch sizes of 32 and 8. All experiments run on NVIDIA H800-80G GPUs.

\textbf{Benchmarks.} We evaluate our method on multi-image grounding benchmarks MIG-Bench~\cite{li2025migician}, MC-Bench~\cite{xu2024mc}, and ReVOS~\cite{yan2024visa}, along with the object detection dataset ODINW~\cite{li2022grounded} and single-image grounding datasets LISA~\cite{lai2024lisa} and Ref-L4~\cite{chen2025revisiting}. To further assess the generalizability of our RL training, we additionally evaluate on two multi-image understanding benchmarks: BLINK~\cite{fu2024blink} and MMIU~\cite{meng2024mmiu}.

\textbf{Baselines.} For comparison, we select several open-source MLLMs, including Qwen2.5-VL~\cite{bai2025qwen2}, InternVL3~\cite{zhu2025internvl3}, LLaVA-OneVision~\cite{li2024llava}, MiniCPM2.6~\cite{yao2024minicpm}, and mPLUG-Owl3~\cite{ye2024mplug}. We further compare our method with the SFT baseline to verify the effectiveness of our RL training.

\textbf{Evaluation Metric.} We adopt Acc@0.5 for grounding evaluation, where a prediction is correct if its Intersection over Union (IoU) with the ground truth exceeds 0.5.

\begin{table}[t]
\centering
\resizebox{0.48\textwidth}{!}{
\begin{tabular}{l|cccc|c|c|c}
\toprule

\multirow{2}{*}{\textbf{Ablation}}  & \multicolumn{4}{c|}{\textbf{Setting}} & \textbf{Referential} & \textbf{Spontaneous} & \multirow{2}{*}{\textbf{AVG}} \\
\cmidrule(lr){2-5}

& \textbf{CS} & \textbf{RS} & \textbf{RL} & $\boldsymbol{\beta}$ & \textbf{Grounding} & \textbf{Grounding} & \\
\midrule

Qwen2.5-VL-7B & - & - & - & - & 10.97 & 18.39 & 13.19 \\
\midrule
w/o CS & \usym{2717} & \usym{2717} & \usym{2713} & \textbf{1e-3} & 14.35 & 25.81 & 17.79 \\
CoT-SFT (Full) & \usym{2713} & \usym{2717} & \usym{2717} & - & 45.34 & 56.92 & 48.82 \\
CoT-SFT (LoRA) & \usym{2713} & \usym{2717} & \usym{2717} & - & 50.62 & 56.56 & 52.41 \\
w/o RS & \usym{2713} & \usym{2717} & \usym{2713} & \textbf{1e-3} & 53.95 & 58.12 & 55.20 \\

w/o KL & \usym{2713} & \usym{2713} & \usym{2713} & \textbf{0.0} & 54.20 & 57.95 & 55.33 \\

\rowcolor{lightblue} \textbf{Ours-Full} & \usym{2713} & \usym{2713} & \usym{2713} & \textbf{1e-3} & 58.16 & 59.21 & 58.47 \\
\bottomrule
\end{tabular}
}
\vspace{-0.3cm}
\caption{\textbf{Ablations on MIG-Bench}. CS, RS, RL are represented as cold-start, rejection sampling and reinforcement learning, respectively. }
\label{tab:ablation_simple}
\vspace{-0.5cm}
\end{table}

\vspace{-0.3cm}
\subsection{Comparison Results} \label{sec:4_2_comparsion}
\vspace{-0.1cm}

As shown in Table~\ref{tab:mig-bench}, our method achieves state-of-the-art performance on the multi-image grounding benchmark MIG-Bench~\cite{li2025migician}. Specifically, it achieves the best results across all seven subsets of referential grounding. Compared with the base model (Qwen2.5-VL-7B), our approach improves performance by 45.64\% and surpasses the second-best model (Qwen2.5-VL-72B) by 12.60\%, while using significantly fewer parameters.

To comprehensively evaluate the robustness and generalization of our method, we assess it from three key perspectives: zero-shot grounding, object detection, and multi-image understanding. As shown in Table~\ref{tab:all_bench}, our full model achieves the highest overall average score of 58.23\%, outperforming Qwen2.5-VL-7B by 12.13\% and the SFT baseline by 4.41\%. 

\textbf{Zero-shot Grounding.} We conduct zero-shot evaluation on diverse datasets, including multi-image grounding (MC-Bench~\cite{xu2024mc}, ReVOS~\cite{yan2024visa}) and single-image grounding (LISA~\cite{lai2024lisa}, Ref-L4~\cite{chen2025revisiting}). Our method achieves an average improvement of 5.06\% over the SFT baseline.

\textbf{Object Detection.} On ODINW~\cite{li2022grounded}, our model achieves the highest average score of 54.15\%, outperforming Qwen2.5-VL-7B by 7.24\%. This indicates robust performance under distribution shifts in out-of-domain detection benchmarks.

\textbf{Multi-image Understanding.} Our RL-enhanced training achieves state-of-the-art performance, surpassing the SFT baseline by an average of 3.17\% on two multi-image understanding datasets, demonstrating that our strategy strengthens grounding while improving higher-level reasoning capabilities.

\vspace{-0.3cm}
\subsection{Ablation Studies} \label{sec:4_3_ablation}
\vspace{-0.1cm}

We perform ablations on MIG-Bench~\cite{li2025migician}, as shown in Table~\ref{tab:ablation_simple}.

\textbf{Effectiveness of Cold Start.} 
As discussed in Section~\ref{sec:3_1_cold_start}, we first apply GRPO-based RL training with the DeepSeek-R1-Zero framework~\cite{guo2025deepseek} (Table~\ref{tab:ablation_simple}, second row), yielding only modest gains (+3.74\%) due to the base model’s limited capacity for multi-image contexts and complex queries. With cold-start initialization (third and fourth rows), performance improves, with LoRA slightly outperforming full-parameter tuning (+0.59\%).

\textbf{Effectiveness of Rejection Sampling.}  To assess the impact of rejection sampling, we train the Stage-1 CoT-SFT model with GRPO but without rejection sampling. As shown in the fifth row of Table~\ref{tab:ablation_simple}, enabling rejection sampling improves performance by 3.27\%, highlighting the importance of data quality and our filtering process.

\textbf{Ablations for KL Divergence.} As shown in the sixth row of Table~\ref{tab:ablation_simple}, removing KL regularization in Stage-2 RL training reduces performance by 3.14\%.
\vspace{-0.4cm}
\section{Conclusion}
\vspace{-0.2cm}

In this paper, we aim to enhance the capabilities of MLLMs for real-world multi-image grounding. We propose a post-training strategy that combines cold-start initialization with rule-based reinforcement learning. Experimental results show that our approach substantially outperforms both the base model and the SFT baseline across multiple grounding and understanding benchmarks. We hope this work will encourage further research in multi-image reasoning and grounding.
\bibliographystyle{IEEEbib}
\bibliography{ref}

\end{document}